# A Study on Mitigating Hard Boundaries of Decision-Tree-based Uncertainty Estimates for AI Models

**Pascal Gerber, Lisa Jöckel, Michael Kläs**
Fraunhofer Institute for Experimental Software Engineering IESE,
Fraunhofer Platz 1, 67663 Kaiserslautern, Germany
{pascal.gerber, lisa.joeckel, michael.klaes}@iese.fraunhofer.de

### Abstract

Outcomes of data-driven AI models cannot be assumed to be always correct. To estimate the uncertainty in these outcomes, the uncertainty wrapper framework has been proposed, which considers uncertainties related to model fit, input quality, and scope compliance. Uncertainty wrappers use a decision tree approach to cluster input quality related uncertainties, assigning inputs strictly to distinct uncertainty clusters. Hence, a slight variation in only one feature may lead to a cluster assignment with a significantly different uncertainty. Our objective is to replace this with an approach that mitigates hard decision boundaries of these assignments while preserving interpretability, runtime complexity, and prediction performance. Five approaches were selected as candidates and integrated into the uncertainty wrapper framework. For the evaluation based on the Brier score, datasets for a pedestrian detection use case were generated using the CARLA simulator and YOLOv3. All integrated approaches achieved a softening, i.e., smoothing, of uncertainty estimation. Yet, compared to decision trees, they are not so easy to interpret and have higher runtime complexity. Moreover, some components of the Brier score impaired while others improved. Most promising regarding the Brier score were random forests. In conclusion, softening hard decision tree boundaries appears to be a trade-off decision.

## 1. Introduction

An increasing number of software and software-intensive systems contain data-driven components, i.e., components that implement functionality using data-driven models (DDMs) as provided, e.g., by Machine Learning (ML) and other AI methods. However, due to their nature of being empirically defined on data, DDMs will not provide correct results in any application situation (Kläs 2018). Therefore, *uncertainty* is an inherence concept of data-driven components, which needs to be considered to make informed decisions. Specifically, this means uncertainty needs to be quantified and either dealt with on the system level or appropriately provisioned to decision makers who use the system.

From a design perspective, two options exist for providing situation-aware uncertainty estimates: (a) The DDMs itself is made responsible for providing not only its primary outcome, but also an estimate on the uncertainty in its outcome, or (b) there is an independent instance that estimates the uncertainty in the DDM outcome (Jöckel and Kläs 2021). In the first case, which is the more deeply researched setting, we talk about *in-model* uncertainty estimation; in the second case, which realizes 'separation of concerns' as a design principle, we talk about *outside-model* uncertainty estimation. *Uncertainty wrappers* (UWs), as an outside-model approach, have recently shown to provide advantages compared to state-of-the-art in-model approaches, such as a higher degree of interpretability and statistically sound uncertainty estimates (Jöckel and Kläs 2021).

*Research Problem* – In order to achieve high interpretability, UWs rely on learning a decision tree structure considering semantic – i.e., human-interpretable – factors influencing the uncertainty (Jöckel and Kläs 2021) (Kläs and Jöckel, 2020) (Kläs et al. 2021). Although this simple structure with clear decision boundaries supports interpretability, it also causes situations in which a minor change in just one factor can result in a significant change in the estimated uncertainty (cf. Fig. 1). While this behavior does not invalidate the results from a statistical point of view, it can be unintuitive in some cases.

For example, in the case of camera-based pedestrian detection, the distance between the camera and the pedestrian could be one such factor, where a lack of smooth transitions in uncertainty is a rather undesirable property of decision trees, as it is for other factors with linear rather than discontinuous effects (Molnar, 2019).

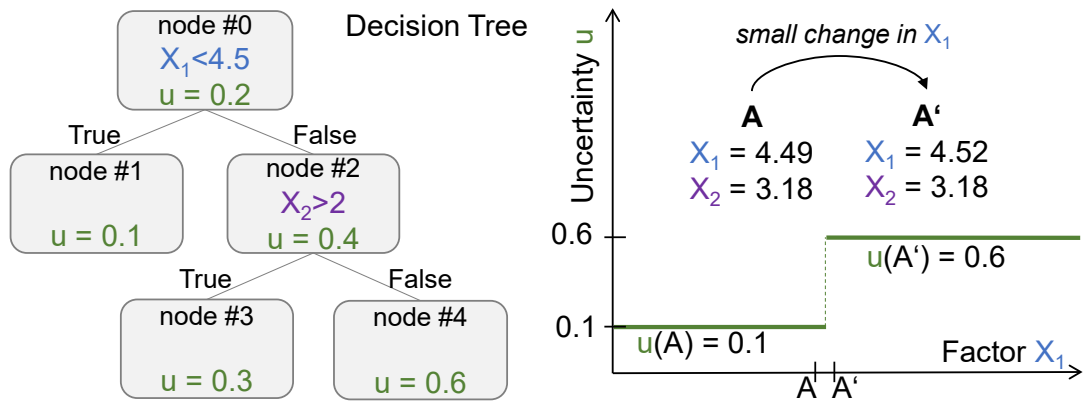

**Fig. 1.** Example of how a minor change in one input factor ($X_1$) may affect the uncertainty estimate (u).

Intuitively, uncertainty does not change significantly as distance only changes marginally (e.g., increases from 14.9m to 15.1m). Instead, it is expected to increase gradually with rising distance, which also improves robustness against noisy inputs.

*Research Goals – To* address this undesired property of decision-tree-based uncertainty estimates, we identified and studied a selection of promising approaches to softening the transitions between different uncertainty levels. Our selection and evaluation, which we present in this paper, was guided by the goals of (G1) making the transitions between different levels of uncertainty smoother while (G2) reducing interpretability as little as possible, (G3) not increasing runtime complexity too much, and (G4) not negatively affecting the uncertainty estimation performance.

*Outline* – Section 2 gives an overview of related work on uncertainty estimation including UWs and possible approaches to softening decision boundaries. Section 3 introduces the investigated approaches and discusses implications on G1 to G3. Section 4 outlines the design and execution of a study to investigate the approaches. Section 5 presents the study results. Section 6 discusses the achievement of the research goals and Section 7 concludes the paper.

## 2. Related Work

***Uncertainty Estimation.*** Uncertainty is an active research topic in the field of ML (Hüllermeier and Waegemann 2019). Getting a better understanding of the sources of uncertainty and providing dependable uncertainty estimations regarding how much we can rely on a specific DDM-based outcome contributes to enabling quality assurance of ML-based systems, especially in contexts with high dependability demands. Uncertainty predictions can also be used to address the fulfillment of system-level performance constraints, such as safety constraints, when using DDMs for some of the system functionality (Kläs et al. 2021).

A classification for potential sources of uncertainty is proposed by the onion shell model (Kläs and Vollmer, 2018), which allows mathematically separating uncertainty due to model fit, input quality, and scope compliance (Kläs and Sembach 2019). Model-fit-related uncertainty is caused by limitations in the DDM itself and can be reduced by improving the model regarding typical performance metrics (e.g., mean absolute error or true positive rate). Input-quality-related uncertainty is concerned with the influence of the quality of the DDM inputs on the quality of the DDM outcomes. Scope-compliance-related uncertainty is caused by possible application of the model outside its application scope and can be tackled by monitoring its compliance.

Uncertainty predictions can be obtained by different means. Some DDMs already implicitly provide uncertainty predictions, e.g., decision trees (Breiman et al. 2017). Other ML approaches have been extended accordingly; for example, for deep neural networks Bayesian neural networks (Arnez et al. 2020) and deep ensembles (Arnez et al. 2020) (Lakshminarayanan, Pritzel, and Blundell 2017) are commonly proposed. Both approaches are, however, computationally expensive and are generally not dependable in the sense that their predictions have no statistical guarantees.

***Uncertainty Wrapper.*** An alternative to uncertainty predictions provided by the DDM itself is the UW framework, which addresses the uncertainty sources of the onion shell model following the separation-of-concerns principle (Kläs and Jöckel, 2020). UWs are model-agnostic and encapsulate the DDM as a black box (cf. Fig. 2).

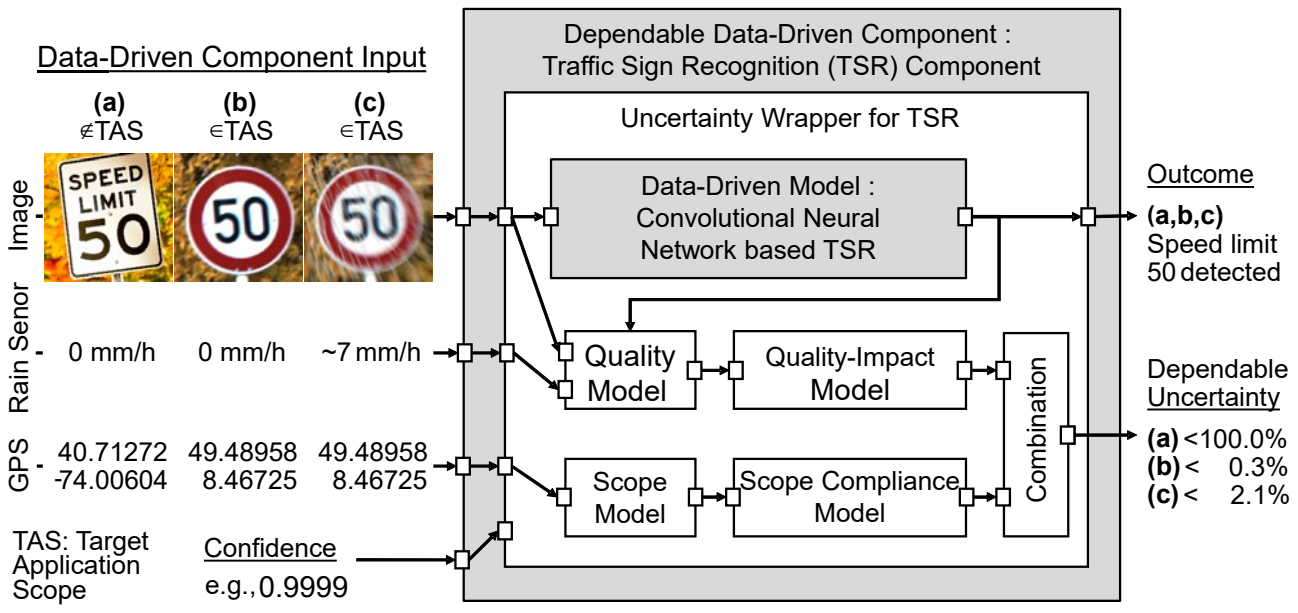

**Fig. 2.** Uncertainty wrapper pattern (Bandyszak et al. 2021).

Jöckel and Kläs (2021) benchmarked the performance of UWs to uncertainty predictions provided by deep ensembles and examined the usefulness of each approach under different non-ideal conditions with which DDMs are confronted in real settings, such as a limited training data, concluding that UWs are beneficial especially in such situations.

Model-fit-related uncertainty can be determined by using model testing approaches and appropriate performance metrics. In order to determine the likelihood of not being in the *target application scope* (TAS), i.e., suffering from scope incompliance, the UW contains a *scope model* and a *scope compliance model*. In the context of traffic sign recognition, the scope model might consider the GPS coordinates of the vehicle as a scope factor that is checked for compliance with a specific input to boundaries defined as the TAS. Similarly, input-quality-related uncertainty is estimated using a *quality model* and a *quality impact model*. For the quality model, quality factors are considered that might occur within the TAS, like obstructed vision due to rain or a dirty camera lens. The quality model determines the presence of each quality factor based on one or more measures, e.g., a rain sensor or a convolutional neural network trained to detect rain (cf. Fig. 2). This information is then used as independent variables together with the correctness of the DDM outcome as a dependent variable in a decision-tree-based quality impact model to find clusters that decompose the TAS into areas with similar uncertainties. An example of such a decision tree together with the associated uncertainties determined for the identified clusters based on the number of correct and wrong DDM outcomes can be seen in Fig.3.

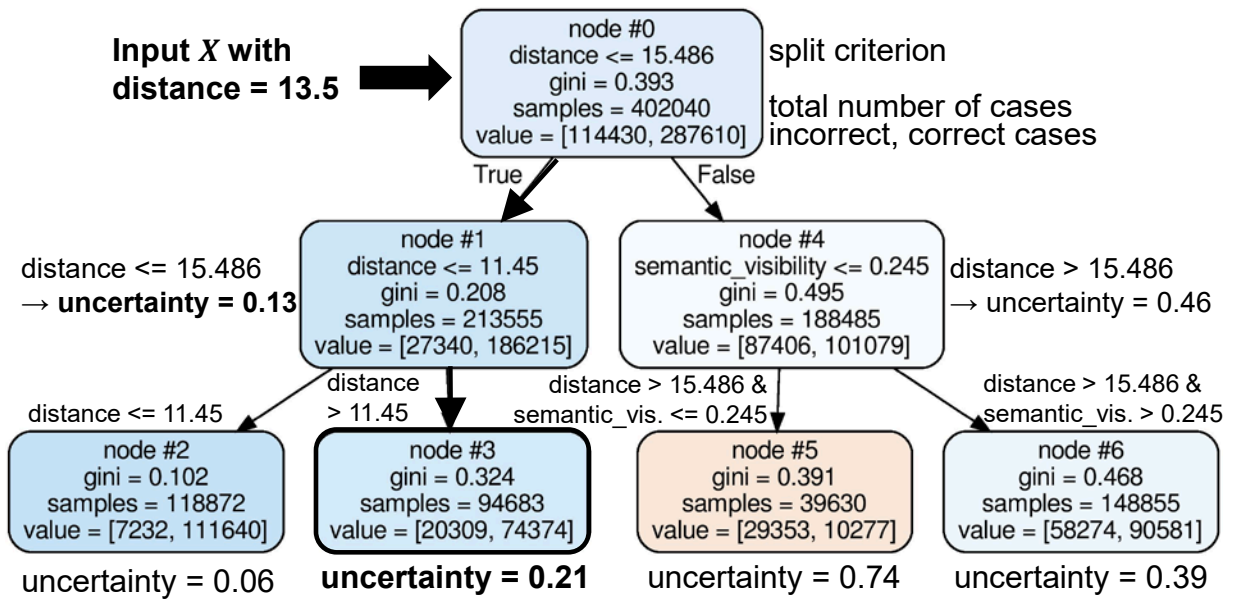

**Fig. 3.** Decision tree as part of a simplified quality impact model.

The uncertainty estimates for the clusters, i.e., the tree nodes, are calibrated using a dataset that has to be representative for the TAS and not previously used to learn the decision tree structure. The UW combines the uncertainty estimates from the scope compliance model and the quality impact model to provide a situation-aware uncertainty estimate for the current input. As *uncertainty* in the context of UWs is defined as the likelihood that the model outcome is incorrect, a *dependable uncertainty* estimate is a justified upper boundary for a given *confidence level* CL (Kläs and Sembach 2019). Hence, the UW includes the additional functionality of considering a requested CL in the estimates.

### Approaches to Softening Decision Boundaries

To identify and select approaches that comply with the goals defined in *Research Goals*, we first searched on Google Scholar (Scholar 2020). Goals G1 and G2 provided an initial basis for creating the search criterion. For 'softening', which is intended by G1, we included the search terms 'soft', 'smooth', or 'fuzzy'. For G2, addressing interpretability, we added the term 'interpret'. No search term was defined for the runtime complexity requirement defined in G3; instead, the identified approaches were examined in this regard after the search. This resulted in the following search criterion:

('decision tree' OR 'breiman') AND
('soft' OR 'smooth' OR 'fuzzy') AND ('interpret')

It should be noted that in the first block of the search criterion, we included the name of the approach as well as the name of the author who presented the CART algorithm for it (Breiman et al. 1984). With this, we intended to find publications describing approaches that aim to mitigate the problem of hard decision boundaries.

The first approach we identified was the Fuzzy Random Forest (Fuzzy RF) (Marcelloni, Matteis, and Segatori 2015), which is an ensemble of several Fuzzy Decision Trees (Fuzzy DTs). As a second approach, the Bagged Soft Decision Trees (Bagged Soft DTs) (Alpaydin, Irsoy, and Yildiz 2016), an ensemble of Soft Decision Trees (Soft DTs) (Alpaydin, Irsoy, and Yildiz 2012), was identified. In addition to this search, we also examined approaches that exist in scikit-learn (Blondel et al. 2012). Here, the Random Forest (RF), an ensemble consisting of DTs, was identified.

Boosting approaches, e.g., AdaBoost or Gradient Tree Boosting, which are mentioned in (Scikit 2020a), were excluded. In these approaches, a model is built iteratively from several base models whose uncertainty estimates are then weighted for the final uncertainty estimate. We anticipated that interpretation (G2) would be difficult to fulfill due to the different weighting of the base models.

## 3. Studied Approaches

This section describes the selected approaches, i.e., Random Forest (RFs), Fuzzy DTs, Fuzzy RFs, Soft DTs, and Bagged Soft DTs, and discusses their compliance in terms of softening uncertainty estimates (G1), interpretability (G2) and runtime complexity (G3).

**Random Forests.** An RF can be considered as a collection of $n$ DTs that are trained and applied independently and whose individual results are aggregated (e.g., by averaging). In order to obtain an ensemble of sufficiently uncorrelated DTs, the DTs in an RF are commonly trained on datasets generated by randomly drawing datapoint samples with replacement from a single dataset, i.e., bootstrapped datasets. Additionally, in contrast to a regular DT, the split feature of a DT node in an RF is determined from a subset rather than from all data point features (Breiman 2001) (Scikit 2020b).

***Goal compliance.*** Regarding (G1), it is expected that by aggregating, some degree of softening of hard decision boundaries can be achieved. Since DTs are interpretable (G2), RFs can also be interpreted. However, all $n$ DTs of an RF need to be considered, resulting in higher effort than for a single DT. Analogously, runtime complexity (G3) for providing uncertainty estimates increases linearly with the number of DTs used. For application in UWs, this complexity seems still acceptable.

**Fuzzification of Decision Trees and Random Forests.** In DTs, each node contains a split criterion, which is used to strictly assign a data point to the left or the right child node. As illustrated in Fig. 3, these criteria are recursively evaluated with specific feature values of a data point in order to finally assign the data point to a leaf with an associated uncertainty. Contrary to the strict assignments in DTs, there are Fuzzy DTs and Fuzzy RFs (Marcelloni, Matteis, and Segatori 2015) that perform fuzzy assignments of data points to nodes on the basis of fuzzy logic or fuzzy set theory (Zadeh 1965). Here, fuzzy partitioning is used to determine a numerical membership degree of a data point to the child nodes of a parent node.

In Fig. 4, an example Fuzzy DT and the fuzzy partitioning of its root node, i.e., node 0, are depicted. Here, the membership degree of a data point in the child nodes of the root is derived by the feature ‘distance’ and the three fuzzy set membership functions, which are uniquely defined by the three values listed below the feature.

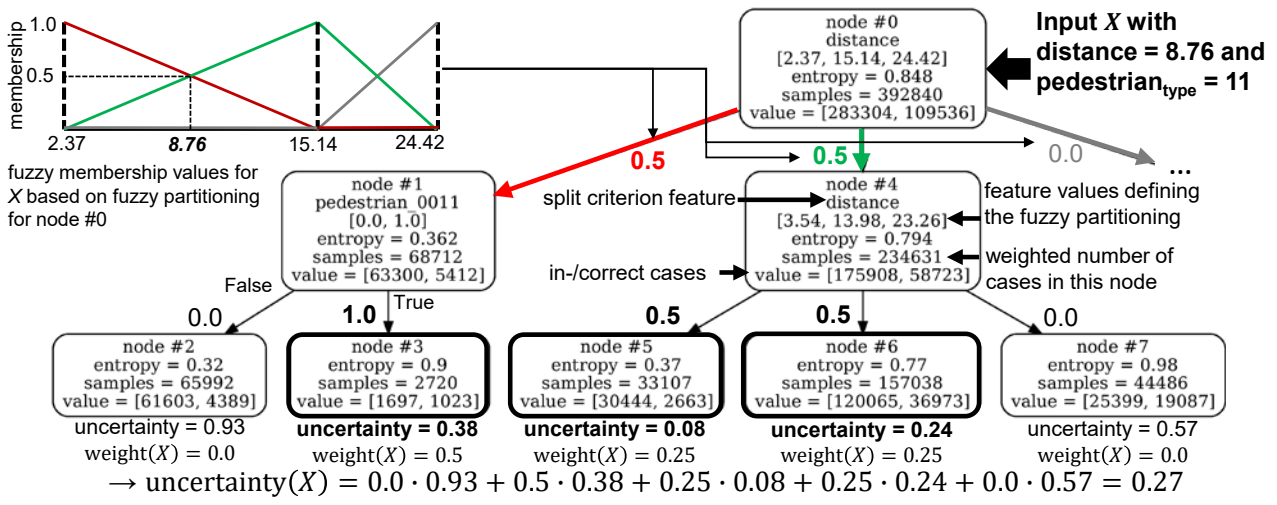

**Fig. 4.** Application of a Fuzzy DT on a data point.

The fuzzy set membership value of a data point with a value of 8.76 for the feature ‘distance’ in node 1 and node 4 is 0.5 and in node 8 it is 0, which means that the data point is further propagated to the left and middle sub-tree with a weight of 0.5 each. This is done recursively towards the leaves of the tree. In order to determine an uncertainty estimate, a weighted sum is built by the uncertainties associated with each leaf, which is weighted by the propagated fuzzy set membership values on the path towards the leaf. Analogously, for a Fuzzy RF, the mean of all uncertainty estimates of the contained Fuzzy DTs is calculated as the final uncertainty estimate.

Using features of a continuous data type as split criterion, fuzzy partitioning is based on three fuzzy sets as illustrated for node 0 in Fig. 4. For categorical features, Marcelloni, Matteis, and Segatori (2015) proposed using crisp sets for each category. But this creates as many child nodes as there are categories of the feature used for splitting, this makes global interpretability of the model more difficult because some of the categories might not be meaningful differentiators between uncertainty clusters. To avoid this, we use one-hot encoding for the categorical features, i.e., each category of the feature is considered as a binary stand-alone feature, leading to a split into two mutually exclusive child nodes, with one being assigned weight 1 and the other weight 0.

A Fuzzy DT is created similarly to the CART algorithm for DTs, but taking into account fuzzy membership partitioning. Starting from the root node, each feature is considered as a splitting criterion in order to determine a partitioning. In doing so, the partitioning is intended to reduce the impurity of the data point labels and can be quantified by the fuzzy information gain based on (fuzzy) entropies for the node to be partitioned and its child nodes. Both of these numbers are depicted in the nodes of Fig. 4 as ‘entropy’. For categorical features, the splitting is done similarly to DTs, as we have only two child nodes. For continuous features, on the other hand, a fuzzy partitioning that is defined by three ascending floating-point numbers (cf. Fig. 4) is created. The first and last number are defined by the minimum and maximum of the occurring feature values. The second number is chosen in between with respect to the fuzzy information gain for all remaining feature values. In order to reduce computational effort, we modified the original approach in our implementation to use equal-width binning on the feature values and only computed the partitioning for one representative of each bin.

***Goal compliance.*** Given that the membership degree of a data point to nodes in Fuzzy DTs and Fuzzy RFs is determined by linear membership functions, the resulting weighting of leaves is not significantly affected by small changes of the feature values. A softening of the uncertainty estimates (G1) is thereby achieved.

With regard to interpretability (G2), a Fuzzy DT can be interpreted on a global level in a similar way as a traditional DT, as the feature and the respective feature values that define the partitioning can be examined (as depicted by the values in brackets in Fig. 4). However, as one data point in a Fuzzy DT is usually a member of two child nodes during partitioning, in the worst case $2^h$ leaves have to be considered for a Fuzzy DT of height $h$ in order to interpret the uncertainty estimate for a concrete input. This is significantly more than in the case of a DT, but interpretability is still feasible assuming a height roughly between 5 and 8. Analogously to RFs, the $n$ contained Fuzzy DTs must also be interpreted additionally in order to interpret Fuzzy RFs.

Similar to interpretability, the runtime complexity (G3) for providing uncertainty estimates increases exponentially with the height of a Fuzzy DT. For Fuzzy RFs, the number of Fuzzy DTs used has an additional linear influence. This may limit the applicability of Fuzzy DTs in certain settings with high Fuzzy DT height and tight runtime requirements.

**(Bagged) Soft Decision Trees.** Alpaydin, Irsoy, and Yildiz introduced Soft DTs (2012) and Bagged Soft DTs, an ensemble of Soft DTs (2016). Similarly to Fuzzy DTs, a data point is propagated from a node to its child nodes with different weights. Whereas for Fuzzy DTs, the weights are derived from membership functions of a fuzzy partitioning, the weights in Soft DTs are determined by soft decisions. For a node $m$, these are based on a sigmoid function $g_m$, which is parameterized by $w_m$, as depicted in Fig. 5.

To determine an uncertainty estimate for a data point, the weighting in each node is determined based on the sigmoid function, and the data point is propagated throughout the tree considering the weights (i.e., similarly to Fuzzy DTs).

The resulting uncertainty estimate associated with an input is, as in the case of Fuzzy DTs, determined by a weighted average over the leaves considering the multiplied weights along each path towards the leaf and the uncertainty associated with the leaf. Analogously, for a Bagged Soft DT, the mean of all uncertainty estimates of the contained Soft DTs is calculated as the final uncertainty estimate.

Unlike in the case of Fuzzy DTs, all features of a data point are used to determine the weighting. However, as this strongly impedes interpretability, we modified our implementation to only use the feature that contributes most to the split in terms of entropy, while the others are ignored as split criterion for this node; i.e., we use a univariate split as for Fuzzy DTs and DTs. Invariant to the approach as described by Alpaydin, Irsoy, and Yildiz (2012), the parameter $w_m$ of the sigmoid function in a node $m$ is determined for this feature using the gradient descent method. In Fig. 5, the respective feature is depicted in the second row of each inner node.

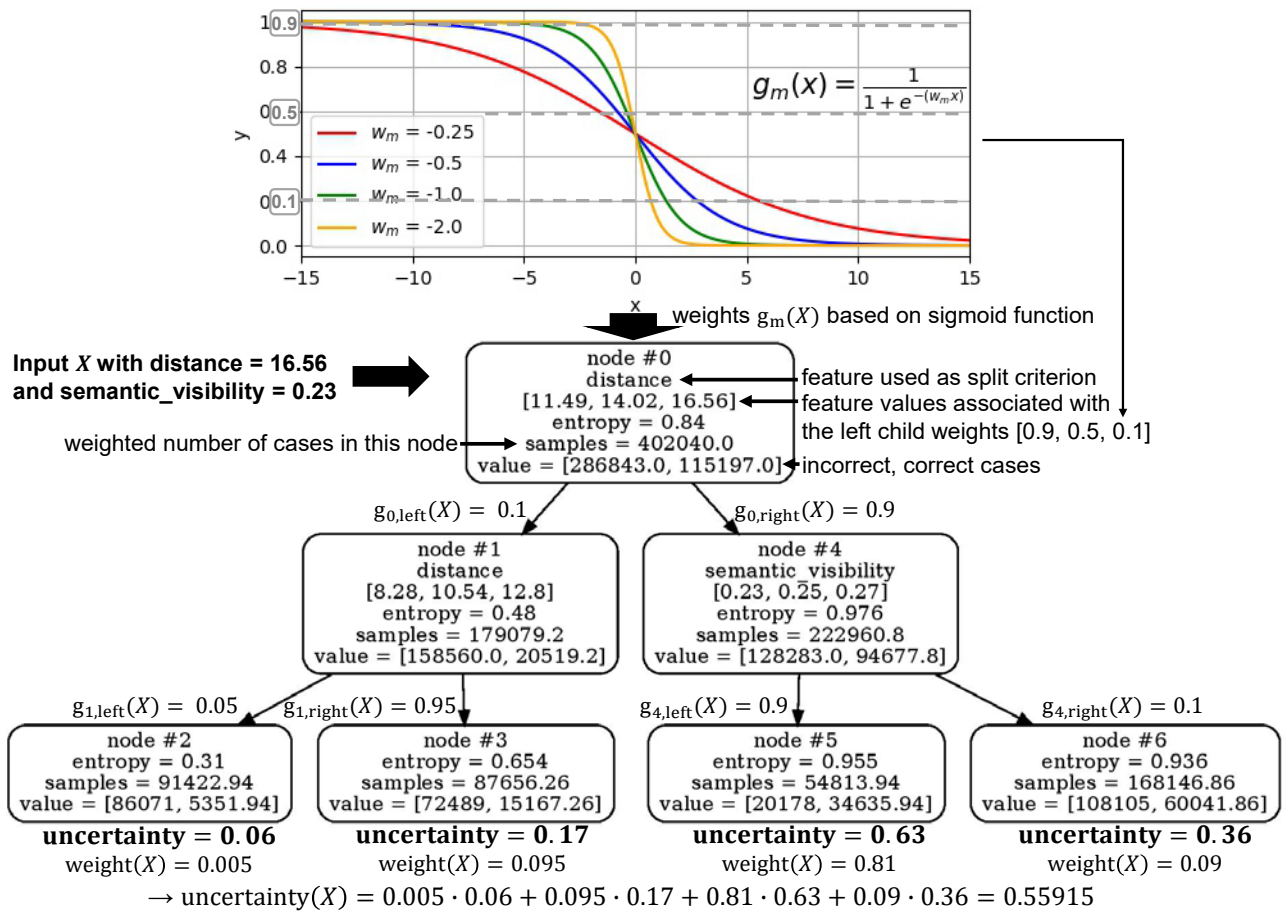

**Fig. 5.** Application of a soft decision tree on a data point.

***Goal compliance.*** Softening of uncertainty estimates (G1) in Soft DTs and Bagged Soft DTs mainly depends on the parameter $w_m$, which drives the steepness of the continuous sigmoid function in a tree node. For a very steep sigmoid function, this softening behavior converges to that of a regular DT. Else, small changes of the sigmoid function argument do not result in significant changes of the resulting weighting, which enables achieving our goal of softening.

The considerations regarding interpretability (G2) are similar to Fuzzy DTs, i.e., the feature used as split criterion is defined for each node based on one feature, and an idea of what the weighting looks like over the feature range is provided by the list of the three feature values in the tree visualization. For a concrete input, the weights can also be tracked while traversing the tree, similarly to Fuzzy DTs. Once again, the effort increases exponentially with the height of a Soft DT, as in the worst case, all paths to the leaves need to be considered. For Bagged Soft DTs, all individual trees need to be evaluated.

Analogously, the runtime complexity (G3) increases exponentially with the height of a Soft DT; for Bagged Soft DTs, it also depends linearly on the number of Soft DTs.

# 4. Study Planning and Execution

In this section, we concretize research goal G4 with the specific research question that we addressed in the study, and present the derived study design and its execution.

**Research Question.** The main research question is how the use of softening approaches affects the performance of uncertainty estimation. For this purpose, we wanted to evaluate the decision-tree-based approach that has been used so far in the UW, as well as Random Forests (RFs), Fuzzy DTs, Fuzzy RFs, Soft DTs, and Bagged Soft DTs.

> *Research Question:* How does the uncertainty estimation performance of the UW differ when softening approaches are used instead of a decision-tree-based approach?

**Task and Target Application Scope.** In order to investigate our research question, we considered the use case of pedestrian detection. In this context, the task of a DDM is to correctly detect pedestrians who are present in given camera images. As the target application scope, we defined a vehicle traveling at various points in time in an urban environment located in Germany. Additionally, we considered situations of pedestrians located up to a maximum distance of 25 meters from the vehicle under a wide variety of weather conditions. Our intention was to provide several factors that influence the input data quality of the images and thus possibly the correctness of the DDM outcomes.

**Study Data.** In the study, we used three datasets: a training dataset to create quality impact models based on the traditional DT and the softening approaches; a calibration dataset, and an evaluation dataset representative for the TAS to calibrate, respectively evaluate, the uncertainty estimates of the quality impact models.

**Study Execution.** Next, we outline the study execution describing the key activities and decisions in the seven steps introduced by the study execution plan (cf. Fig. 6).

**(1) Generate raw data.** The datasets were generated using the open-source simulator CARLA (Codevilla et al. 2017), that allows visually and physically realistic simulation of vehicles, pedestrians, and environmental conditions.

For the recording, a vehicle was automatically navigated through a town with other traffic participants like vehicles, trucks, motorcycles, bicycles, and pedestrians. The images of pedestrians crossing the road in the driving direction were recorded, along with measures related to their respective region in the image, the distance, the level of occlusion, and the type of pedestrian. Also, the weather parameters simulated for a sequence of images, such as sun position, cloudiness, precipitation, fog/wind intensity, and road wetness were recorded. While weather parameters were *uniformly* varied for the recordings related to the training dataset, the ones related to the calibration/evaluation datasets were intended to be *representative* for the target application scope.

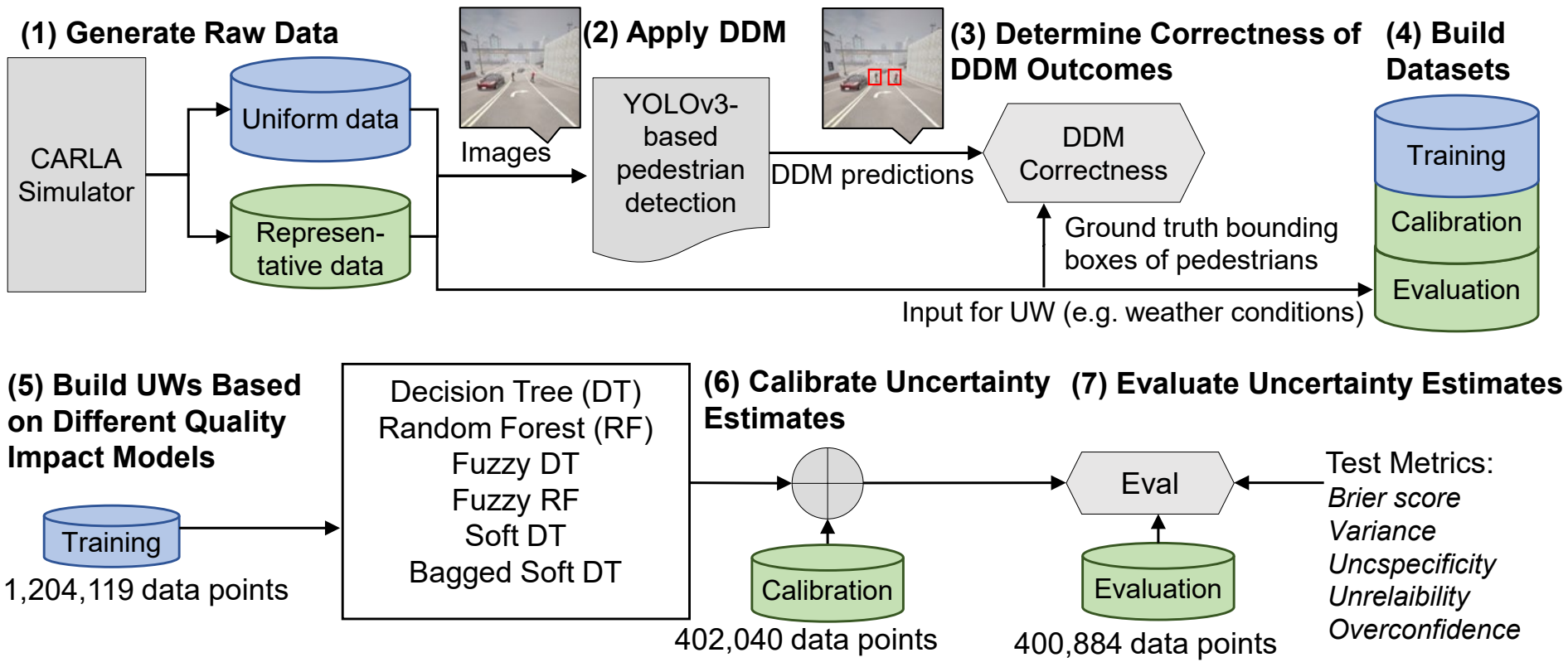

**Fig. 6.** Summarized study execution plan with execution steps.

Hence, weather observations from CDC (2020) over a period of ten years were randomly sampled and weather parameters were derived for the CARLA simulator. As indicated in Fig. 6, we saved the respective raw data in two datasets. Fig. 7 provides some samples of the different weather conditions and locations that were simulated in CARLA.

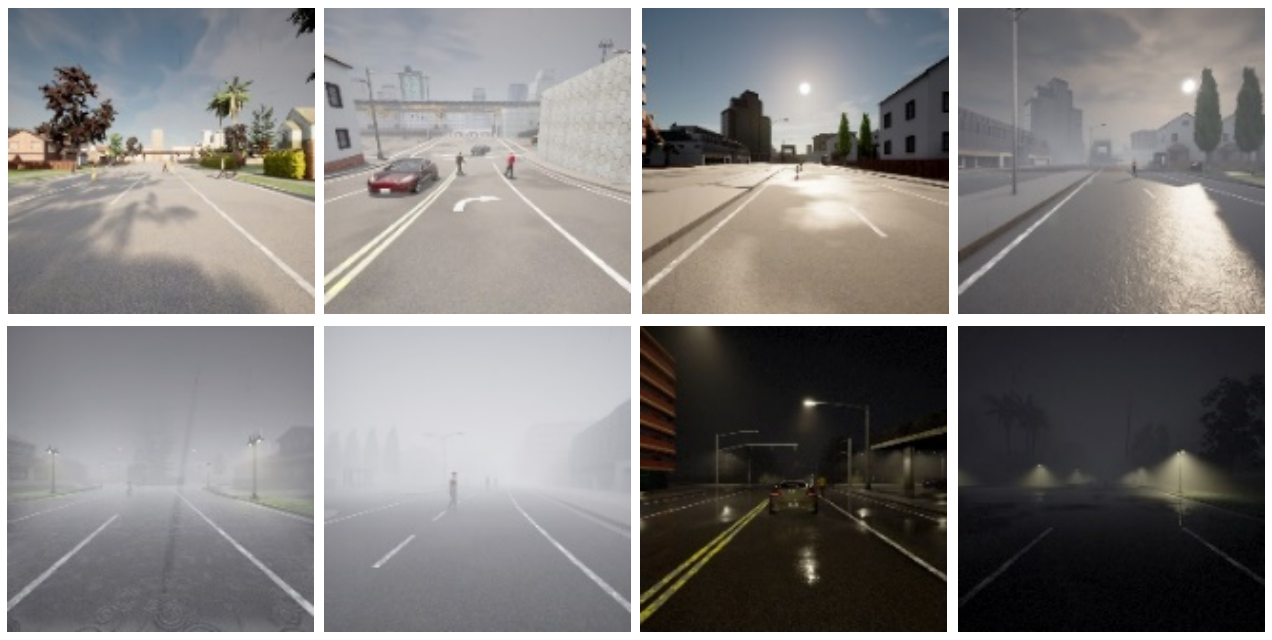

**Fig. 7.** An excerpt of the CARLA simulator's weather conditions.

**(2) Apply DDM.** To obtain the pedestrian detection outcomes for the images saved in the two datasets, we used the YOLOv3 implementation (YOLO 2020) as the encapsulated DDM. For this, pretrained network weights exist that have been learned on the COCO (2020) dataset, that contains a class for persons. Prior to creating the outcomes, the object confidence threshold ("conf thres") was set to 0.5 and the non-maximum suppression threshold ("nms thres") was set to 0.4. Next, outcomes defining enclosing rectangles for each object detected were obtained for each image. On this basis, YOLOv3 created bounding boxes with a class label as an outcome for each object detected in an image. Since we investigated the performance of the uncertainty wrapper and not of the DDM, we did not apply a sophisticated hyperparameter search or other efforts to optimize the DDM.

**(3) Determine correctness of DDM outcomes.** The next step was to determine whether the pedestrians in the image were correctly detected by YOLOv3. To decide this, the intersection over union metric (Gwak et al. 2019), also referred to as Jaccard index, was first used to quantify how well a bounding box detection matches the recorded ground truth bounding box of a pedestrian. Edge cases in which the ground truth bounding boxes of pedestrians were less than 50% visible within the image were filtered out. We defined that outcomes resulting in intersection over union values of at least 0.25 were considered to be correct and lower values were considered to be incorrect. As a result, the correctness of the YOLOv3 outcome was determined and added to each recorded image of both raw datasets.

**(4) Build datasets.** Based on both datasets, we built *training*, *calibration,* and *evaluation* datasets, each consisting of the quality model inputs (e.g., weather conditions) and the correctness of the DDM outcomes. The training dataset was based on the uniform data, whereas the calibration and evaluation datasets were based on the representative data. Aiming to assure that no data points of the same recorded simulation sequence were present in the calibration and the evaluation datasets, we performed a randomized split on the sequences. Thus, the training, calibration, and evaluation datasets consisted of 1,204,119, 402,040, and 400,884 data points, respectively.

**(5) Build UWs.** Using the training dataset, the different quality impact model approaches were trained. Here, we manually searched for suitable hyperparameter ranges for each approach, which were then instantiated for training.

**(6) Calibrate uncertainty estimates.** After the training, several quality impact models were obtained for each approach. Aiming to make the quality impact models' uncertainty estimates reliable for the target application scope, calibration was carried out on the calibration dataset considering a confidence level of 99.99%.

**(7) Evaluate uncertainty estimates.** Estimating uncertainty, defined as the probability that a specific DDM outcome is incorrect, can be seen as a binary probabilistic classification task. Therefore, strictly proper scoring rules such as the Brier score (Brier 1950), can be applied for their evaluation. The Brier score measures the mean squared difference between the predicted probability of a DDM outcome and the actual outcome. Besides the overall Brier score ($bs$), we also measured the three additive components into which it can be decomposed (Murphy 1973), namely variance ($var$), resolution ($res$), and unreliability ($unr$):

$$bs = var - res + unr$$

$var$ describes the overall variation in the correctness of the DDM outcomes, which is only dependent on the overall uncertainty of the DDM outcomes and thus not influenced by the uncertainty estimation approach used. For $res$, the difference between the case-specific uncertainty estimates and the overall uncertainty is considered. As higher $res$ values are better and $var$ is the upper bound, we used $var - res$ as *unspecificity* ($uns$), where smaller values mean that the resolution of the estimator is able to detail more of the variance in the uncertainty of the DDM outcomes. $unr$ measures the degree of calibration of the estimator to the observed correctness of the DDM, such that better calibration leads to a smaller $unr$ value. Furthermore, we used the part of the $unr$ value that is attributed to underestimating the observed error rate of the DDM as a metric of *overconfidence* ($oconf$), as this part of unreliability is especially relevant in the context of safety-related functions.

For the evaluation of the DT and the five candidate approaches, we selected the best quality impact model for each approach subject to the $bs$. If there were quality impact models that resulted in approximately equal $bs$ values, the $unr$ values were also considered and the quality impact model with the lowest $unr$ value was selected. The idea was that a low $unr$ value implies high reliability of the quality impact model's uncertainty estimates.

## 5. Study Results

In this section, we will answer the research question on the basis of our study results. To answer how the performance of uncertainty estimation differs when using alternative softening approaches instead of the previous DT-based quality impact model, we present the evaluation results in Table 1.

**Table 1.** Evaluation results per quality impact model approach.

| Approach | $bs$ | $var$ | $uns$ | $unr$ | $oconf$ |
|---|---|---|---|---|---|
| DT | .12176 | .20284 | .12109 | .00066 | 7.4e-07 |
| RF | .11908 | .20284 | .00646 | .11261 | .05918 |
| Fuzzy DT | .13568 | .20284 | .01252 | .12316 | .07617 |
| Fuzzy RF | .13799 | .20284 | .00407 | .13393 | .08611 |
| Soft DT | .16239 | .20284 | .00001 | .16238 | .10664 |
| Bagged Soft DT | .16279 | .20284 | .00001 | .16277 | .10751 |

Considering the overall performance measured by the Brier score ($bs$), RF performed best, while DT performed slightly worse. This was followed by the fuzzy approaches, where Fuzzy RF performed slightly worse than Fuzzy DT. Soft DT and Bagged Soft DT, which had approximately the same $bs$, represent the worst results.

For DT, the $unr$ term has a very low value, and the value of $bs$ is essentially constituted by the $uns$ term. For RF, this is reversed, i.e., $bs$ is primarily constituted by the $unr$ term. Similar results can be seen with Fuzzy RF and Fuzzy DT, with the latter scoring even higher $uns$. Soft DT and Bagged Soft DT show minor differences in terms of $uns$ and $unr$. Here, the high values of $bs$ are mainly caused by the $unr$ term. In addition, compared to DT, all approaches show an increased $oconf$ and even a bias toward overconfident uncertainty estimates.

## 6. Discussion

Compared to DTs, we were able to achieve a softening of uncertainty estimates (G1) with all other approaches. As an example, we illustrate this in Fig. 8 by means of the uncertainty estimates for an example data point, which we variate in a single quality factor. To provide a better visual overview, the uncertainty estimates of Fuzzy DT and Soft DT are not displayed, as they demonstrated similar characteristics as those of the respective ensemble approaches for this combination of feature values. While the uncertainty estimates of DTs vary considerably at the decision boundaries, the uncertainty estimates of RF vary less. Both the Fuzzy and Soft approaches show even softer boundaries.

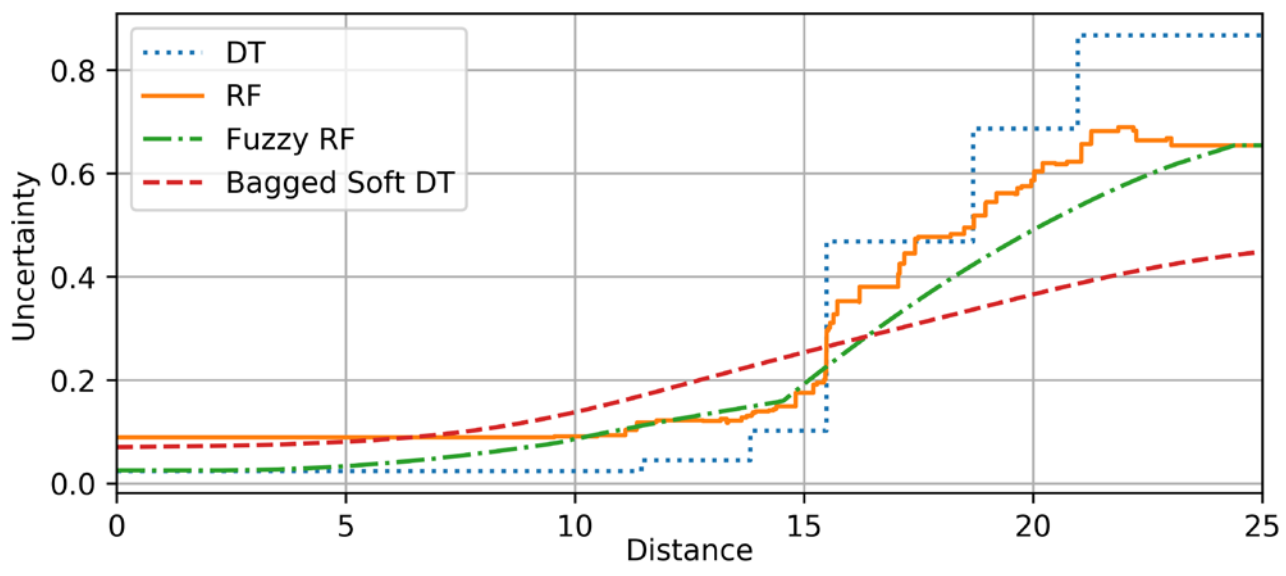

**Fig. 8.** Estimated uncertainties of a sample data point.

The interpretability (G2) of the models and of the concrete uncertainty estimates for data points is feasible for all approaches, which is why we assessed the effort required for this. For DTs, the effort was the lowest, since the split criteria ensure that there is always only one path to a leaf. For RFs, this effort increased with the number of DTs used, which also applied to the other ensemble approaches. For Fuzzy and Soft approaches, there are usually several paths to leaves that are weighted differently, which increases the effort for interpretation exponentially with the tree height.

Analogously to the effort for interpreting, the runtime complexity (G3) can be assessed. For DTs, the tree height has a linear influence. In the case of RF, the number of DTs contained has an additional linear influence. In contrast, for the Fuzzy and Soft approaches, the depth of a tree has a significant influence on this complexity, so the applicability of these approaches may be limited in some settings with high tree height and tight runtime requirements.

The use of softening approaches results in a significantly lower unspecificity because of their higher resolution compared to DTs. However, unreliability increases in exchange, with the result that the overall uncertainty estimation performance (G4) is worse for most of the investigated approaches in comparison to that of DTs. Although RF slightly outperform DTs in overall uncertainty scores, their significantly worse unreliability and overconfidence scores make them less attractive to use in many cases. A summary of the extent to which the approaches achieve our goals is given in Table 2. There, for each goal an assessment is made on a scale ranging from '--' to '++', where the former represents poor and the latter very good achievement of the goal.

**Table 2.** Goal fruition rating per quality impact model approach.

| Approach | G1 Soft. | G2 Interpret. | G3 Run. complex. | G4 Est. perform. |
|---|---|---|---|---|
| DT | -- | ++ | ++ | ++ |
| RF | o | + | + | + |
| Fuzzy DT | ++ | + | o | o |
| Fuzzy RF | ++ | o | - | o |
| Soft DT | ++ | + | o | - |
| Bagged Soft DT | ++ | o | - | - |

## 7. Conclusion

We investigated alternative approaches to decision trees that allow softening of uncertainty estimates while preserving reasonable interpretability, runtime complexity, and uncertainty estimation performance. In an empirical study, we compared the uncertainty estimation performance of these approaches using the example task of pedestrian detection and the Brier score with its subcomponents as a metric.

The results do not allow providing a general recommendation for the use of a particular approach; rather, the selection of an approach has proved to be a trade-off decision (cf. Table 2), which should be made considering the planned application. Random forests, closely followed by decision trees, showed the best results considering their overall uncertainty estimation performance but at the cost of significantly higher unreliability as decision trees. Fuzzy decision trees and fuzzy random forests as well as soft decision trees and bagged soft decision trees showed excellent performance in softening the decision boundaries but reduced uncertainty estimation performance. In general, all investigated softening approaches showed a lower but still acceptable level of interpretability and runtime performance compared to decision trees as our baseline. However, in specific settings with tight runtime requirements or high demands on interpretability, the application of certain approaches may be limited.

For features with known or expected discontinuities, the use of hard decision boundaries rather than soft transitions may seem more appropriate. Therefore, it should be noted that all approaches allow features of categorical and continuous data type, where the softening is only carried out for the continuous features. This means that all features for which hard decision boundaries are more appropriate should be designed as categorical data in a preprocessing step.

Based on the study results, we see two main directions for further work on softening the boundaries of decision-tree-based uncertainty estimation. First, specific recommendations should be developed that guide practitioners in deciding in which settings they should make use of which of the investigated approaches. Second, the investigated approaches could be further modified to address the observed limitations regarding uncertainty estimation performance, specifically addressing unreliability and overconfidence.

## Acknowledgments

Parts of this work have been funded by the Observatory for Artificial Intelligence in Work and Society (KIO) of the Denkfabrik Digitale Arbeitsgesellschaft in the project "KI Testing & Auditing", the Federal Ministry for Economic Affairs and Energy in the project "SPELL", and by the project "LOPAAS" as part of the internal funding program "ICON" of the Fraunhofer-Gesellschaft. We would like to thank Sonnhild Namingha for the initial review of the paper.